\def\year{2020}\relax
\documentclass[letterpaper]{article} 
\usepackage{aaai20}  
\usepackage{amsmath}
\usepackage{times}  
\usepackage{helvet} 
\usepackage{courier}  
\usepackage[hyphens]{url}  
\usepackage{graphicx} 
\usepackage{graphicx}  
\usepackage{multirow}
\usepackage{amssymb}
\usepackage{subcaption}
\usepackage{booktabs,tabularx}
\newcommand{\minisection}[1]{\noindent{\textbf{#1}}}
\title{Real-time Scene Text Detection with Differentiable Binarization}
\author{
Minghui Liao\textsuperscript{\rm 1}\thanks{Authors contribute equally.},
Zhaoyi Wan\textsuperscript{\rm 2}\footnotemark[\value{footnote}],
Cong Yao\textsuperscript{\rm 2},
Kai Chen\textsuperscript{\rm 3,4},
Xiang Bai\textsuperscript{\rm 1}\thanks{Corresponding author}\\
\textsuperscript{\rm 1}Huazhong University of Science and Technology,
\textsuperscript{\rm 2}Megvii,
\textsuperscript{\rm 3}Shanghai Jiao Tong University,
\textsuperscript{\rm 4}Onlyou Tech.\\
\{mhliao,xbai\}@hust.edu.cn, i@wanzy.me, yaocong2010@gmail.com, kchen@sjtu.edu.cn
}
\begin{document}

\maketitle

\begin{abstract}
Recently, segmentation-based methods are quite popular in scene text detection, as the segmentation results can more accurately describe scene text of various shapes such as curve text. However, the post-processing of binarization is essential for segmentation-based detection, which converts probability maps produced by a segmentation method into bounding boxes/regions of text. 
In this paper, we propose a module named Differentiable Binarization (DB), which can perform the binarization process in a segmentation network. Optimized along with a DB module, a segmentation network can adaptively set the thresholds for binarization, which not only simplifies the post-processing but also enhances the performance of text detection. Based on a simple segmentation network, we validate the performance improvements of DB on five benchmark datasets, which consistently achieves state-of-the-art results, in terms of both detection accuracy and speed.
In particular, with a light-weight backbone, the performance improvements by DB are significant so that we can look for an ideal tradeoff between detection accuracy and efficiency.
Specifically, with a backbone of ResNet-18, our detector achieves an F-measure of 82.8, running at 62 FPS, on the MSRA-TD500 dataset. Code is available at: https://github.com/MhLiao/DB.
\end{abstract}

\section{Introduction}

In recent years, reading text in scene images has become an active research area, due to its wide practical applications such as image/video understanding, visual search, automatic driving, and blind auxiliary. 

As a key component of scene text reading, scene text detection that aims to localize the bounding box or region of each text instance is still a challenging task, since scene text is often with various scales and shapes, including horizontal, multi-oriented and curved text. Segmentation-based scene text detection has attracted a lot of attention recently, as it can describe the text of various shapes, benefiting from its prediction results at the pixel-level. However, most segmentation-based methods require complex post-processing for grouping the pixel-level prediction results into detected text instances, resulting in a considerable time cost in the inference procedure. Take two recent state-of-the-art methods for scene text detection as examples: PSENet~\cite{wang2019shape} proposed the post-processing of progressive scale expansion for improving the detection accuracies; Pixel embedding in \cite{tian2019learning} is used for clustering the pixels based on the segmentation results, which has to calculate the feature distances among pixels.   

\begin{figure}[tbp]
\centering
\includegraphics[width=0.9\linewidth]{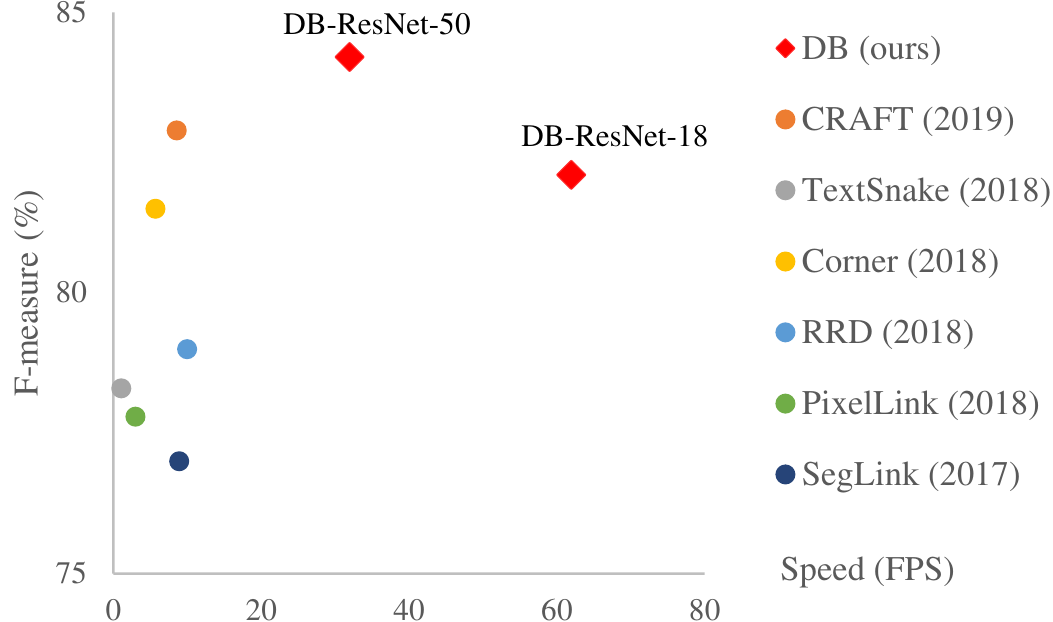}
\caption{The comparisons of several recent scene text detection methods on the MSRA-TD500 dataset, in terms of both accuracy and speed. Our method achieves the ideal tradeoff between effectiveness and efficiency. }
\label{fig:performance_speed}
\end{figure}

\begin{figure}[tbp]
\centering
\includegraphics[width=0.95\linewidth]{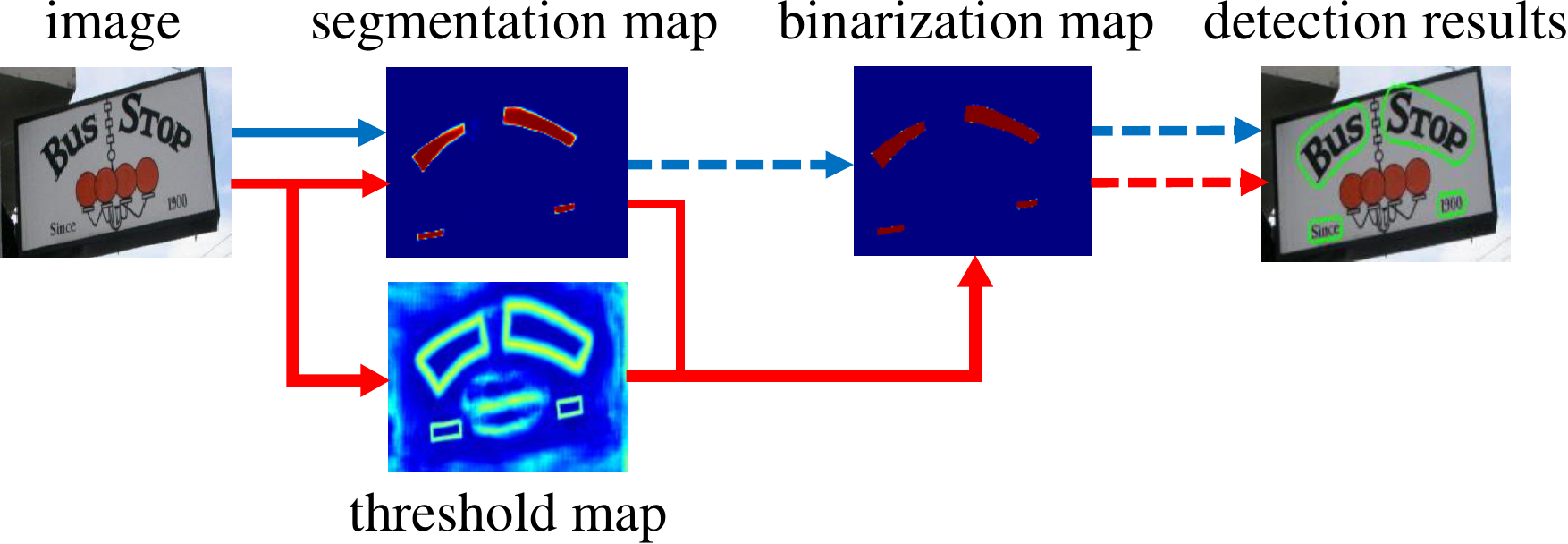}
\caption{Traditional pipeline (blue flow) and our pipeline (red flow). Dashed arrows are the inference only operators; solid arrows indicate differentiable operators in both training and inference.}
\label{fig:intorduction}
\end{figure}

Most existing detection methods use the similar post-processing pipeline as shown in Fig.~\ref{fig:intorduction} (following the blue arrows): Firstly, they set a fixed threshold for converting the probability map produced by a segmentation network into a binary image; Then, some heuristic techniques like pixel clustering are used for grouping pixels into text instances. Alternatively, our pipeline (following the red arrows in Fig.~\ref{fig:intorduction}) aims to insert the binarization operation into a segmentation network for joint optimization. In this manner, the threshold value at every place of an image can be adaptively predicted, which can fully distinguish the pixels from the foreground and background. However, the standard binarization function is not differentiable, we instead present an approximate function for binarization called Differentiable Binarization (DB), which is fully differentiable when training it along with a segmentation network. 

The major contribution in this paper is the proposed DB module that is differentiable, which makes the process of binarization end-to-end trainable in a CNN.
By combining a simple network for semantic segmentation and the proposed DB module, we proposed a robust and fast scene text detector. 
Observed from the performance evaluation of using the DB module, we discover that our detector has several prominent advantages over the previous state-of-the-art segmentation-based approaches.
\begin{enumerate}
    \item Our method achieves consistently better performances on five benchmark datasets of scene text, including horizontal, multi-oriented and curved text.  
    \item Our method performs much faster than the previous leading methods, as DB can provide a highly robust binarization map, significantly simplifying the post-processing.
    \item DB works quite well when using a light-weight backbone, which significantly enhances the detection performance with the backbone of ResNet-18.
    \item As DB can be removed in the inference stage without sacrificing the performance, there is no extra memory/time cost for testing. 
\end{enumerate}

\section{Related Work}
Recent scene text detection methods can be roughly classified into two categories: Regression-based methods and segmentation-based methods.

\minisection{Regression-based methods}
are a series of models which directly regress the bounding boxes of the text instances. TextBoxes~\cite{LiaoSBWL17} modified the anchors and the scale of the convolutional kernels based on SSD~\cite{liu2015ssd} for text detection. TextBoxes++~\cite{TextBoxes++} and DMPNet~\cite{deepmatch} applied quadrilaterals regression to detect multi-oriented text. SSTD~\cite{sstd} proposed an attention mechanism to roughly identifies text regions. RRD~\cite{liao2018rotation} decoupled the classification and regression by using rotation-invariant features for classification and rotation-sensitive features for regression, for better effect on multi-oriented and long text instances. EAST~\cite{east} and DeepReg~\cite{deepdirect} are anchor-free methods, which applied pixel-level regression for multi-oriented text instances. SegLink~\cite{seglink} regressed the segment bounding boxes and predicted their links, to deal with long text instances. DeRPN~\cite{xie2019derpn} proposed a dimension-decomposition region proposal network to handle the scale problem in scene text detection.
Regression-based methods usually enjoy simple post-processing algorithms (e.g. non-maximum suppression). However, most of them are limited to represent accurate bounding boxes for irregular shapes, such as curved shapes.

\minisection{Segmentation-based methods} 
usually combine pixel-level prediction and post-processing algorithms to get the bounding boxes. \cite{Zhang_2016_CVPR} detected multi-oriented text by semantic segmentation and MSER-based algorithms. Text border is used in \cite{xue2018accurate} to split the text instances,
Mask TextSpotter~\cite{lyu2018mask,liao2019mask} detected arbitrary-shape text instances in an instance segmentation manner based on Mask R-CNN. PSENet~\cite{wang2019shape} proposed progressive scale expansion by segmenting the text instances with different scale kernel. Pixel embedding is proposed in \cite{tian2019learning} to cluster the pixels from the segmentation results. 
PSENet~\cite{wang2019shape} and SAE~\cite{tian2019learning} proposed new post-processing algorithms for the segmentation results, resulting in lower inference speed. Instead, our method focus on improving the segmentation results by including the binarization process into the training period, without the loss of the inference speed.

\minisection{Fast scene text detection methods}
focus on both the accuracy and the inference speed. TextBoxes~\cite{LiaoSBWL17}, TextBoxes++~\cite{TextBoxes++}, SegLink~\cite{seglink}, and RRD~\cite{liao2018rotation} achieved fast text detection by following the detection architecture of SSD~\cite{liu2015ssd}. EAST~\cite{east} proposed to apply PVANet~\cite{pvanet} to improve its speed. Most of them can not deal with text instances of irregular shapes, such as curved shape.
Compared to the previous fast scene text detectors, our method not only runs faster but also can detect text instances of arbitrary shapes.

\section{Methodology}
\begin{figure*}[tbp]
\centering
\includegraphics[width=0.9\linewidth]{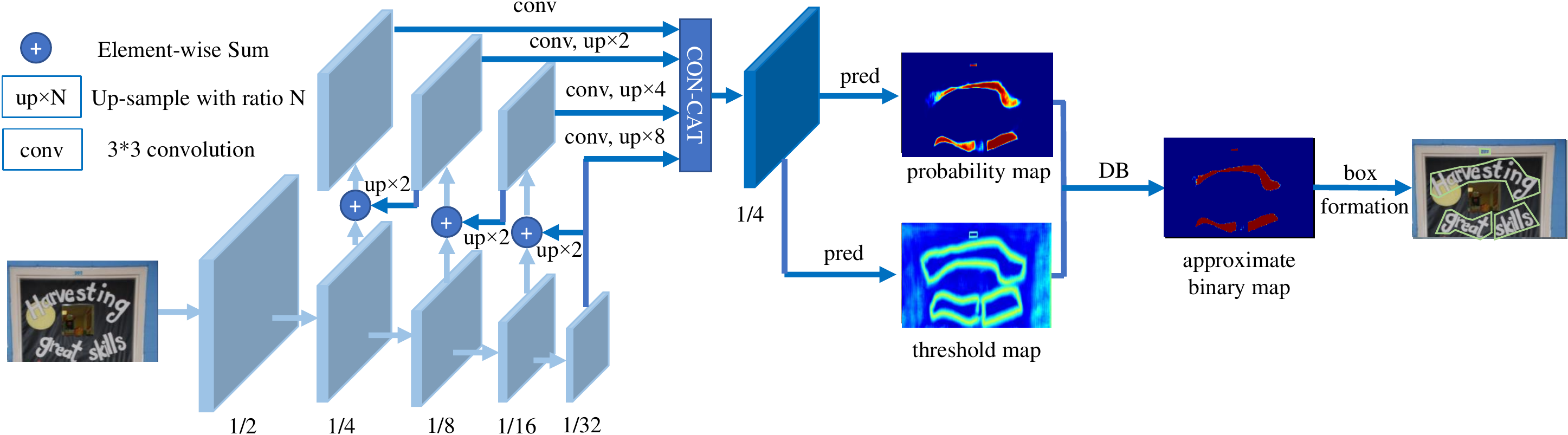}
\caption{Architecture of our proposed method, where ``pred" consists of a $3 \times 3$ convolutional operator and two de-convolutional operators with stride 2. The ``1/2", ``1/4", ... and ``1/32" indicate the scale ratio compared to the input image.}
\label{fig:pipeline}
\end{figure*}

The architecture of our proposed method is shown in Fig.~\ref{fig:pipeline}. Firstly, the input image is fed into a feature-pyramid backbone. Secondly, the pyramid features are up-sampled to the same scale and cascaded to produce feature $F$. Then, feature $F$ is used to predict both the probability map ($P$) and the threshold map ($T$). After that, the approximate binary map ($\hat{B}$) is calculated by $P$ and $F$. In the training period, the supervision is applied on the probability map, the threshold map, and the approximate binary map, where the probability map and the approximate binary map share the same supervision. In the inference period, the bounding boxes can be obtained easily from the approximate binary map or the probability map by a box formulation module. 

\subsection{Binarization}

\begin{figure}[htbp]
\centering
\includegraphics[width=0.9\linewidth]{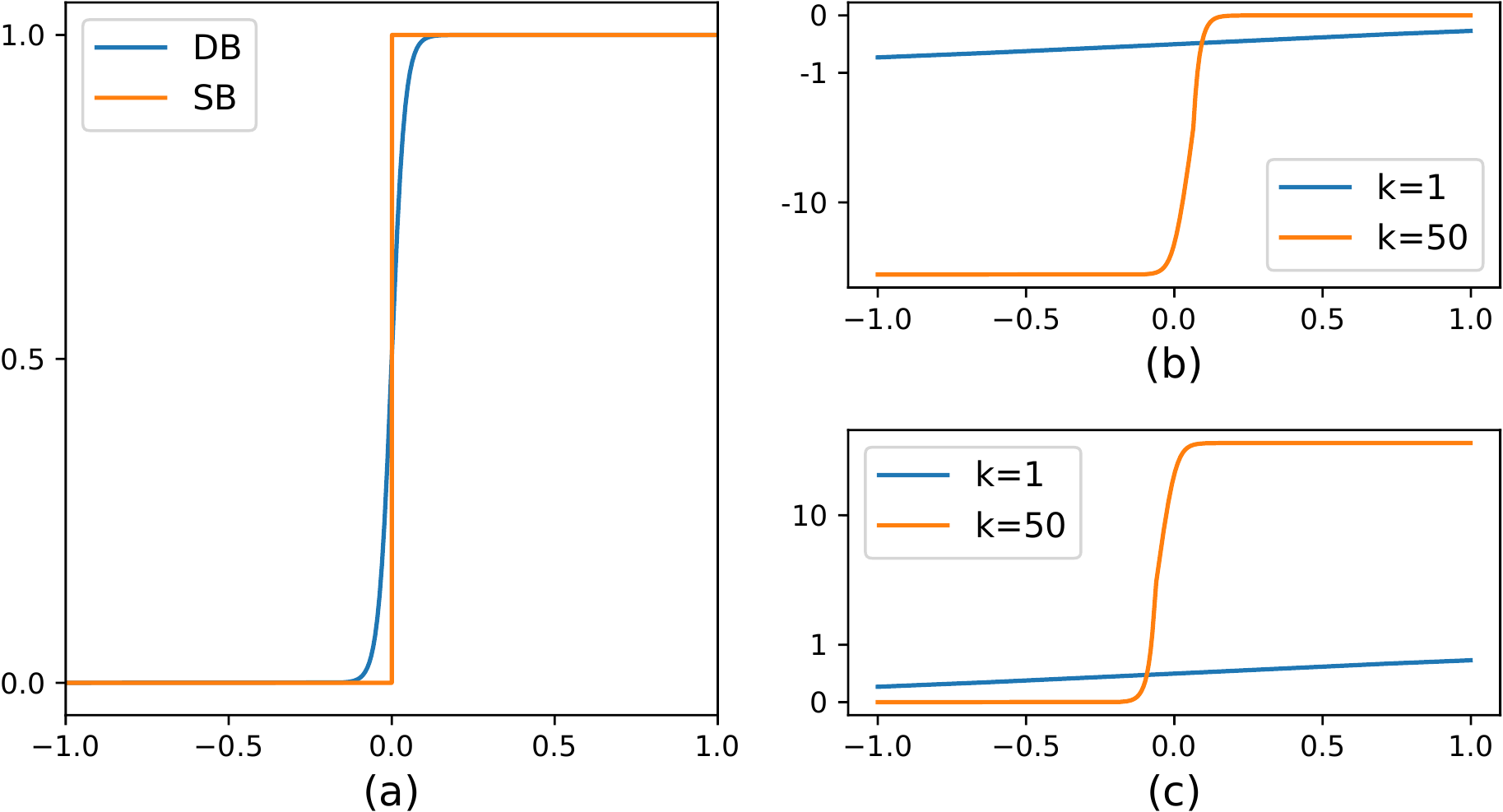}
\caption{Illustration of differentiable binarization and its derivative.
(a) Numerical comparison of standard binarization (SB) and differentiable binarization (DB). (b) Derivative of $l_+$. (c) Derivative of $l_-$.}
\label{fig:step_function}
\end{figure}

\subsubsection{Standard binarization}
Given a probability map $P \in R^{H \times W}$ produced by a segmentation network, where $H$ and $W$ indicate the height and width of the map, it is essential to convert it to a binary map $P \in R^{H \times W}$, where pixels with value $1$ is considered as valid text areas. Usually, this binarization process can be described as follows:
\begin{equation}
    B_{i, j}=
    \begin{cases}
        1& \text{if } P_{i, j} >= t, \\
        0&  \text{otherwise.}
    \end{cases}
    \label{eq:binarization}
\end{equation}
where $t$ is the predefined threshold and $(i, j)$ indicates the coordinate point in the map.

\subsubsection{Differentiable binarization}
The standard binarization described in Eq.~\ref{eq:binarization} is not differentiable. Thus, it can not be optimized along with the segmentation network in the training period. To solve this problem, we propose to perform binarization with an approximate step function:
\begin{equation}
    \hat{B}_{i, j}=\frac{1}{1+e^{-k (P_{i, j} - T_{i, j})}}
    \label{eq:db}
\end{equation}
where $\hat{B}$ is the approximate binary map; $T$ is the adaptive threshold map learned from the network; $k$ indicates the amplifying factor. $k$ is set to $50$ empirically. 
This approximate binarization function behaves similar to the standard binarization function (see Fig~\ref{fig:step_function}) but is differentiable thus can be optimized along with the segmentation network in the training period.
The differentiable binarization with adaptive thresholds can not only help differentiate text regions from the background, but also separate text instances which are closely jointed. Some examples are illustrated in Fig.\ref{fig:visu}.

The reasons that DB improves the performance can be explained by the backpropagation of the gradients. Let’s take the binary cross-entropy loss as an example. Define $f(x) = \frac {1} {1 + e^{-kx}}$ as our DB function, where $x = P_{i, j} - T_{i, j}$. Then the losses $l_{+}$ for positive labels and $l_{-}$ for negative labels are:
\begin{equation}
    \begin{split}
    l_{+} &= - \log \frac {1}{1+e^{-kx}} \\
    l_{-} &= - \log ( 1 - \frac {1}{1+e^{-kx}} )
    \end{split}
\end{equation}
We can easily compute the differential of the losses with the chain rule:
\begin{equation}
    \begin{split}
        \frac {\partial l_{+}}{\partial x} &= -kf(x)e^{-kx}\\
        \frac {\partial l_{-}}{\partial x} &= kf(x)
    \end{split}
\end{equation}
The derivatives of $l_{+}$ and $l_{-}$ are also shown in Fig.~\ref{fig:step_function}.
We can perceive from the differential that (1) The gradient is augmented by the amplifying factor $k$; (2) The amplification of gradient is significant for most of the wrongly predicted region ($x<0$ for $L_{+}$; $x>0$ for $L_{-}$), thus facilitating the optimization and helping to produce more distinctive predictions. Moreover, as $x = P_{i, j} - T_{i, j}$, the gradient of P is effected and rescaled between the foreground and the background by $T$.

\subsection{Adaptive threshold}
\begin{figure}[htbp]
    \begin{subfigure}{0.453\linewidth}
    \centering
        \includegraphics[width=\linewidth]{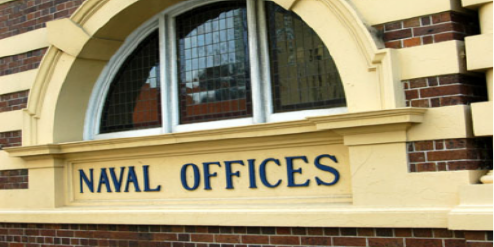}
        \caption{}
    \end{subfigure} %
    \qquad
    \begin{subfigure}{0.453\linewidth}
    \centering
        \includegraphics[width=\linewidth]{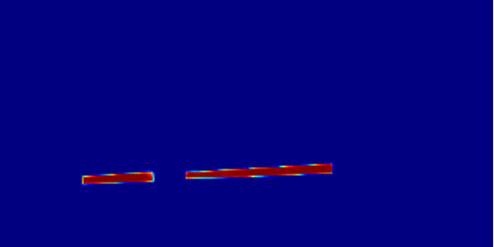}
        \caption{}
   \end{subfigure}

    \begin{subfigure}{0.453\linewidth}
    \centering
        \includegraphics[width=\linewidth]{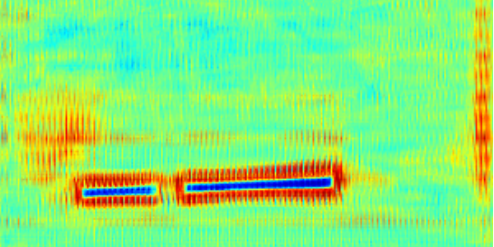}
        \caption{}
    \end{subfigure} %
    \qquad
    \begin{subfigure}{0.453\linewidth}
    \centering
        \includegraphics[width=\linewidth]{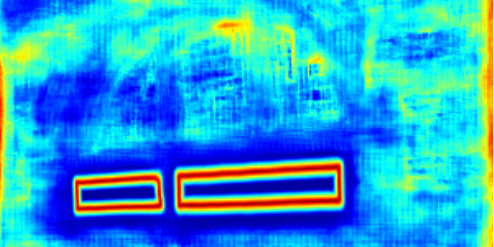}
        \caption{}
   \end{subfigure}
\caption{The threshold map with/without supervision. (a) Input image. (b) Probability map. (c) Threshold map without supervision. (d) Threshold map with supervision.}
\label{fig:supervision}
\end{figure}

The threshold map in Fig.~\ref{fig:performance_speed} is similar to the text border map in \cite{xue2018accurate} from appearance. However, the motivation and usage of the threshold map are different from the text border map. The threshold map with/without supervision is visualized in Fig.~\ref{fig:supervision}. 
The threshold map would highlight the text border region even without supervision for the threshold map. This indicates that the border-like threshold map is beneficial to the final results. Thus, we apply border-like supervision on the threshold map for better guidance. An ablation study about the supervision is discussed in the Experiments section.
For the usage, the text border map in \cite{xue2018accurate} is used to split the text instances while our threshold map is served as thresholds for the binarization. 

\subsection{Deformable convolution}
Deformable convolution~\cite{dai2017deformable,zhu2019deformable} can provide a flexible receptive field for the model, which is especially beneficial to the text instances of extreme aspect ratios. 
Following~\cite{zhu2019deformable}, modulated deformable convolutions are applied in all the $3 \times 3$ convolutional layers in stages conv3, conv4, and conv5 in the ResNet-18 or ResNet-50 backbone~\cite{he2016deep}.

\subsection{Label generation}
\begin{figure}[htbp]
\centering
\includegraphics[width=0.95\linewidth]{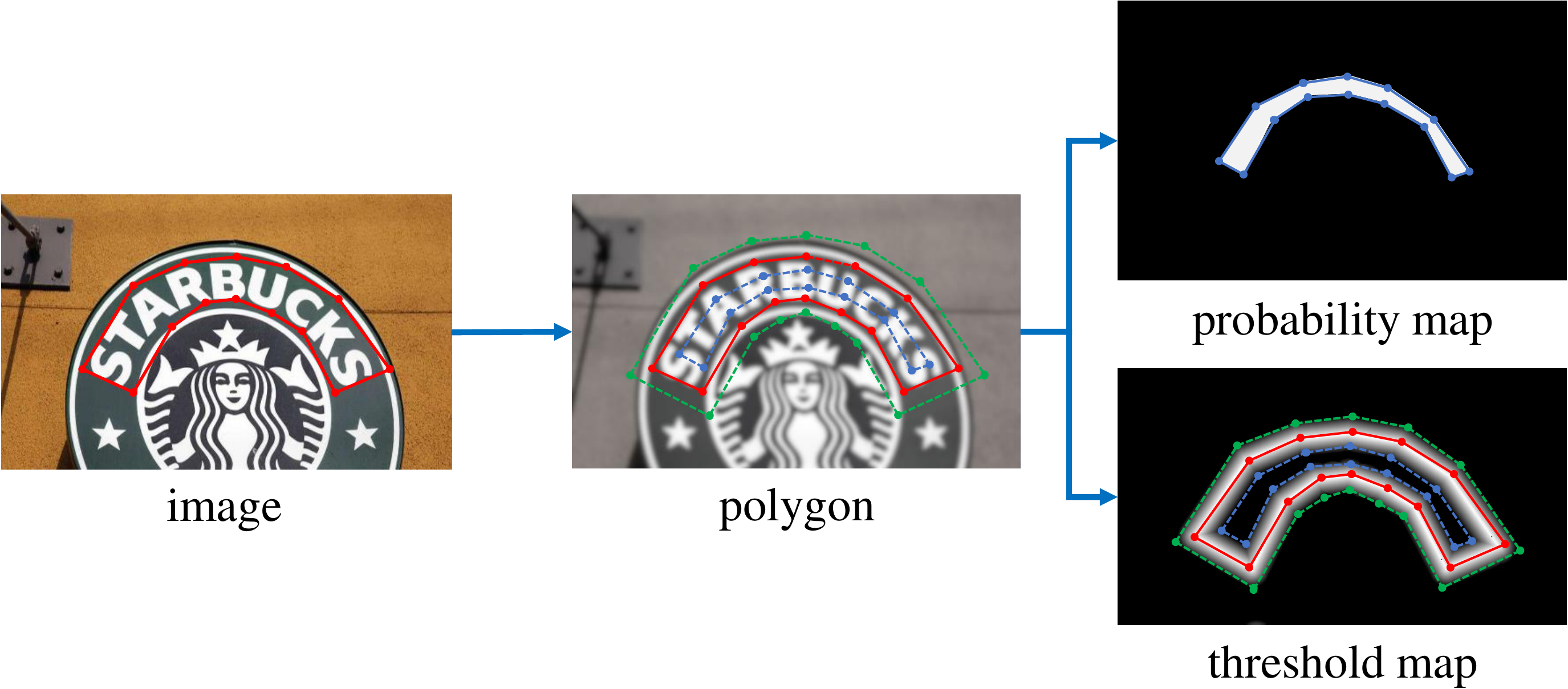}
\caption{Label generation. The annotation of text polygon is visualized in red lines. The shrunk and dilated polygon are displayed in blue and green lines, respectively.}
\label{fig:label}
\end{figure}

The label generation for the probability map is inspired by PSENet~\cite{wang2019shape}. 
Given a text image, each polygon of its text regions is described by a set of segments:
\begin{equation}
    G = \{S_{k}\}^{n}_{k=1}
\end{equation}
$n$ is the number of vertexes, which may be different in different datasets, e.g, 4 for the ICDAR 2015 dataset~\cite{icdar15} and 16 for the CTW1500 dataset~\cite{ctw1500}. Then the positive area is generated by shrinking the polygon $G$ to $G_{s}$ using the Vatti clipping algorithm~\cite{vati}. The offset $D$ of shrinking is computed from the perimeter $L$ and area $A$ of the original polygon: 
\begin{equation}
    D = \frac{A(1-r^{2})}{L}
\end{equation}
where $r$ is the shrink ratio, set to $0.4$ empirically.

With a similar procedure, we can generate labels for the threshold map. Firstly the text polygon $G$ is dilated with the same offset $D$ to $G_{d}$. We consider the gap between $G_{s}$ and $G_{d}$ as the border of the text regions, where the label of the threshold map can be generated by computing the distance to the closest segment in $G$.

\subsection{Optimization}
The loss function $L$ can be expressed as a weighted sum of the loss for the probability map $L_{s}$, the loss for the binary map $L_{b}$, and the loss for the threshold map $L_t$:
\begin{equation}
    L = L_{s} + \alpha \times L_{b} + \beta \times L_{t}
\end{equation}
where $L_s$ is the loss for the probability map and $L_b$ is the loss for the binary map. According to the numeric values of the losses, $\alpha$ and $\beta$ are set to $1.0$ and $10$ respectively.

We apply a binary cross-entropy (BCE) loss for both $L_s$ and $L_b$. To overcome the unbalance of the number of positives and negatives, hard negative mining is used in the BCE loss by sampling the hard negatives.
\begin{equation}
    L_s = L_b = \sum_{i \in S_{l}}{y_i}\log{x_i} + (1-y_i)\log{(1-x_i)}
\end{equation}
$S_{l}$ is the sampled set where the ratio of positives and negatives is $1:3$.

$L_{t}$ is computed as the sum of $L1$ distances between the prediction and label inside the dilated text polygon $G_d$:
\begin{equation}
    L_t = \sum_{i \in R_d}{|y^{*}_i - x^{*}_i|}
\end{equation}
where $R_d$ is a set of indexes of the pixels inside the dilated polygon $G_d$; $y^{*}$ is the label for the threshold map.

In the inference period, we can either use the probability map or the approximate binary map to generate text bounding boxes, which produces almost the same results. For better efficiency, we use the probability map so that the threshold branch can be removed. The box formation process consists of three steps: (1) the probability map/the approximate binary map is firstly binarized with a constant threshold (0.2), to get the binary map; (2)the connected regions (shrunk text regions) are obtained from the binary map; (3) the shrunk regions are dilated with an offset $D'$ the Vatti clipping algorithm\cite{vati}. $D'$ is calculated as 
\begin{equation}
    D' = \frac{A' \times r'}{L'}
\end{equation}
where $A'$ is the area of the shrunk polygon; $L'$ is the perimeter of the shrunk polygon; $r'$ is set to $1.5$ empirically.

\begin{figure*}[htbp]
\centering
\includegraphics[width=0.9\linewidth]{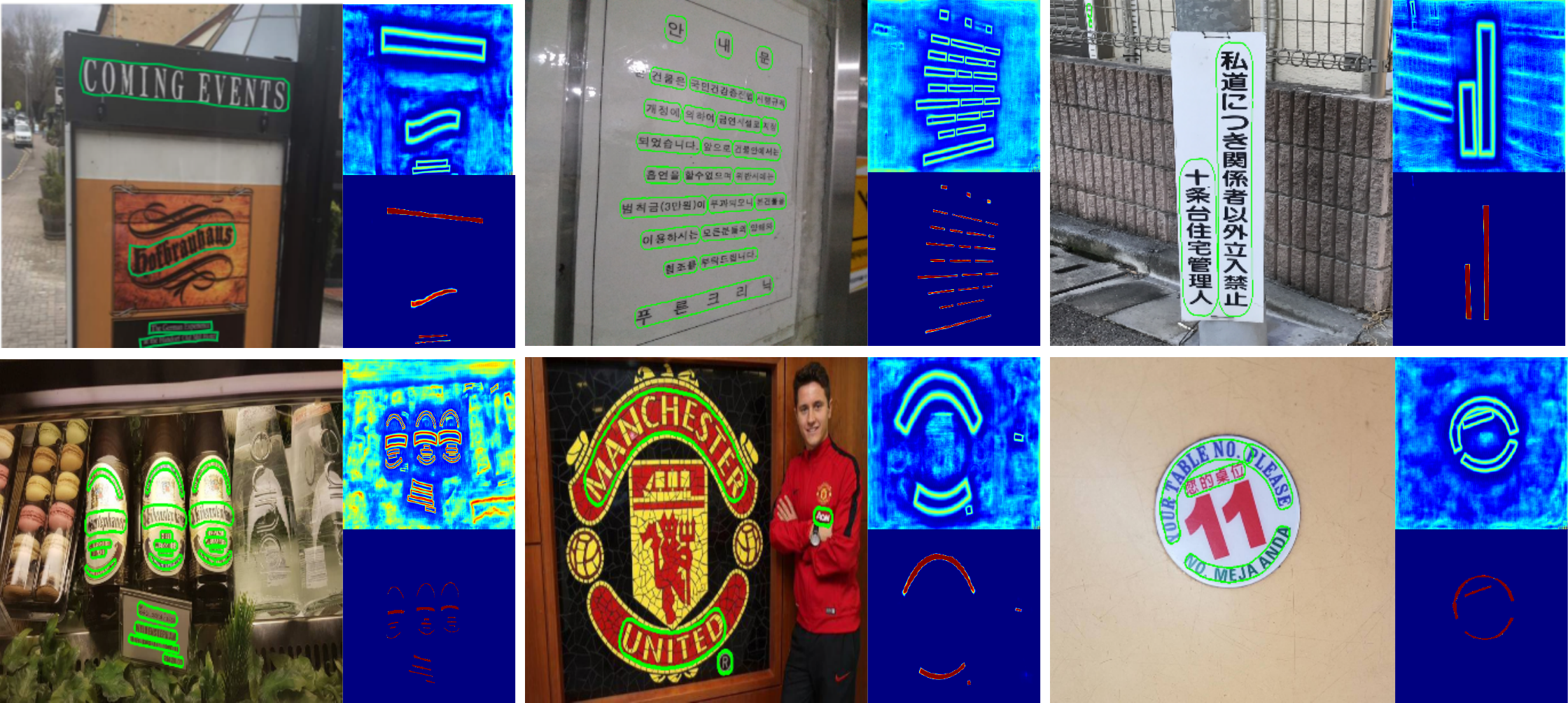}
\caption{Some visualization results on text instances of various shapes, including curved text, multi-oriented text, vertical text, and long text lines. For each unit, the top right is the threshold map; the bottom right is the probability map.}
\label{fig:visu}
\end{figure*}

\begin{table*}[ht]
\centering
\caption{Detection results with different settings. ``DConv" indiates deformable convolution. ``P'', ``R'', and ``F'' indicate precision, recall, and f-measure respectively.}
\begin{tabularx}{1.0\linewidth}{@{}l*{10}X@{}}
\toprule
%
\multirow{2}{*}{Backbone} & \multirow{2}{*}{DConv} & \multirow{2}{*}{DB} & \multicolumn{4}{c}{MSRA-TD500} & \multicolumn{4}{c}{CTW1500} \\ \cline{4-11} 
                          &                                         &                                              & P      & R      & F      & FPS & P      & R      & F      & FPS \\ 
\midrule   
ResNet-18                 & $\times$                                      & $\times$                                           & 85.5   & 70.8   & 77.4   & \textbf{66} & 76.3 & 72.8 & 74.5 & \textbf{59} \\ 
ResNet-18                 & \checkmark                                     & $\times$                                           & 86.8   & 72.3   & 78.9   & 62  & 80.9 & 75.4 & 78.1 & 55 \\ 
ResNet-18                 & $\times$                                      & \checkmark                                          & 87.3   & 75.8   & 81.1   & \textbf{66}  & 82.4 & 76.6 & 79.4 & \textbf{59} \\ 
ResNet-18                 & \checkmark                                     & \checkmark                                          & \textbf{90.4}   & \textbf{76.3}   & \textbf{82.8}   & 62 & \textbf{84.8} & \textbf{77.5} & \textbf{81.0} & 55  \\ \midrule  

ResNet-50                 & $\times$                                      & $\times$                                           & 84.6   & 73.5   & 78.7   & \textbf{40}  & 81.6 & 72.9 & 77.0 & \textbf{27} \\ 
ResNet-50                 & \checkmark                                     & $\times$                                           & 90.5   & 77.9   & 83.7   & 32 & 86.2 & 78.0 & 81.9 & 22 \\ 
ResNet-50                 & $\times$                                      & \checkmark                                          & 86.6   & 77.7   & 81.9   & \textbf{40} & 84.3 & 79.1 & 81.6 & \textbf{27} \\ 
ResNet-50                 & \checkmark                                     & \checkmark                                          & \textbf{91.5}   & \textbf{79.2}   & \textbf{84.9}   & 32 & \textbf{86.9} & \textbf{80.2} & \textbf{83.4} & 22 \\ 
\bottomrule
\end{tabularx}
\label{tab:ablation}
\end{table*}

\section{Experiments}

\subsection{Datasets}
\minisection{SynthText}~\cite{SynthText} is a synthetic dataset which consists of $800k$ images. These images are synthesized from 8k background images. This dataset is only used to pre-train our model.

\minisection{MLT-2017 dataset}~\footnote{https://rrc.cvc.uab.es/?ch=8} is a multi-language dataset. It includes 9 languages representing 6 different scripts. There are 7,200 training images, 1,800 validation images and 9,000 testing images in this dataset. We use both the training set and the validation set in the finetune period. 

\minisection{ICDAR 2015 dataset}~\cite{icdar15} consists of 1000 training images and 500 testing images, which are captured by Google glasses with a resolution of $720 \times 1280$. The text instances are labeled at the word level. 

\minisection{MSRA-TD500 dataset}~\cite{MSRA} is a multi-language dataset that includes English and Chinese. There are 300 training images and 200 testing images. The text instances are labeled in the text-line level. Following the previous methods~\cite{east,lyu2018multi,long2018textsnake}, we include extra 400 training images from HUST-TR400~\cite{yao2014unified}.

\minisection{CTW1500 dataset}
CTW1500~\cite{ctw1500} is a dataset which focuses on the curved text. It consists of 1000 training images and 500 testing images. The text instances are annotated in the text-line level.

\minisection{Total-Text dataset}
Total-Text~\cite{totaltext} is a dataset that includes the text of various shapes, including horizontal, multi-oriented, and curved. They are 1255 training images and 300 testing images. The text instances are labeled at the word level.

\subsection{Implementation details}
For all the models, we first pre-train them with the SynthText dataset for $100k$ iterations. Then, we finetune the models on the corresponding real-world datasets for $1200$ epochs. The training batch size is set to 16. We follow a “poly” learning rate
policy where the learning rate at current iteration equals the initial learning rate
multiplying $(1 - \frac{iter}{max\_iter})^{power}$, where the initial learning rate is set to 0.007 and $power$ is $0.9$. We use a weight decay of 0.0001 and a momentum
of 0.9. The $max\_iter$ means the maximum iterations, which depends on the maximum epochs.

The data augmentation for the training data includes: (1) Random rotation with an angle range of $(-10^{\circ}, 10^{\circ})$; (2) Random cropping; (3) Random Flipping. All the processed images are re-sized to $640 \times 640$ for better training efficiency.

In the inference period, we keep the aspect ratio of the test images and re-size the input images by setting a suitable height for each dataset. The inference speed is tested with a batch size of $1$, with a single 1080ti GPU in a single thread. The inference time cost consists of the model forward time cost and the post-processing time cost. The post-processing time cost is about $30\%$ of the inference time.

\subsection{Ablation study}
We conduct an ablation study on the MSRA-TD500 dataset and the CTW1500 dataset to show the effectiveness of our proposed differentiable binarization, the deformable convolution, and different backbones. The detailed experimental results are shown in Tab.~\ref{tab:ablation}.

\minisection{Differentiable binarization}
In Tab.~\ref{tab:ablation}, we can see that our proposed DB improves the performance significantly for both ResNet-18 and ResNet-50 on the two datasets.
For the ResNet-18 backbone, DB achieves $3.7\%$ and $4.9\%$ performance gain in terms of F-measure on the MSRA-TD500 dataset and the CTW1500 dataset. For the ResNet-50 backbone, DB brings $3.2\%$ (on the MSRA-TD500 dataset)  and $4.6\%$ (on the CTW1500 dataset) improvements. Moreover, since DB can be removed in the inference period, the speed is the same as the one without DB. 

\minisection{Deformable convolution}
As shown in Tab.~\ref{tab:ablation}, the deformable convolution can also brings $1.5-5.0$ performance gain since it provides a flexible receptive field for the backbone, with small extra time costs. For the MSRA-TD500 dataset, the deformable convolution increase the F-measure by $1.5\%$ (with ResNet-18) and $5.0\%$ (with ResNet-50). For the CTW1500 dataset, $3.6\%$ (with ResNet-18) and $4.9\%$ (with ResNet-50) improvements are achieved by the deformable convolution.

\begin{table}[!ht]
\centering
\caption{Effect of supervising the threshold map on the MLT-2017 dataset. ``Thr-Sup'' denotes applying supervision on the threshold map.}
\begin{tabularx}{1.0\linewidth}{@{}ll*{5}X@{}}
\toprule

Backbone & Thr-Sup & P      & R      & F      & FPS  \\               
\midrule                     
ResNet-18                                     & $\times$                                          & 81.3   & 63.1   & 71.0   & 41 \\ 

ResNet-18                                     & \checkmark                                          & \textbf{81.9}   & \textbf{63.8}   & \textbf{71.7}   & 41 \\ \midrule  
ResNet-50                                                    & $\times$                                           & 81.5   & 64.6   & 72.1   & 19   \\ 
ResNet-50                                      & \checkmark                                          & \textbf{83.1}   & \textbf{67.9}   & \textbf{74.7}   & 19 \\ \bottomrule
\end{tabularx}
\label{tab:thresh}
\end{table}

\minisection{Supervision of threshold map}
Although the threshold maps with/without supervision are similar in appearance, the supervision can bring performance gain. As shown in Tab.~\ref{tab:thresh}, the supervision improves $0.7\%$ (ResNet-18) and $2.6\%$ (ResNet-50) on the MLT-2017 dataset.

\minisection{Backbone}
The proposed detector with ResNet-50 backbone achieves better performance than the ResNet-18 but runs slower. Specifically, The best ResNet-50 model outperforms the best ResNet-18 model by $2.1\%$ (on the MSRA-TD500 dataset) and $2.4\%$ (on the CTW1500 dataset), with approximate double time cost.

\subsection{Comparisons with previous methods}
We compare our proposed method with previous methods on five standard benchmarks, including two benchmarks for curved text, one benchmark for multi-oriented text, and two multi-language benchmarks for long text lines. Some qualitative results are visualized in Fig.~\ref{fig:visu}.

\begin{table}[!ht]
\centering
\caption{Detection results on the Total-Text dataset. The values in the bracket mean the height of the input images. ``*" indicates testing with multiple scales. ``MTS'' and ``PSE'' are short for Mask TextSpotter and PSENet.}
\begin{tabularx}{1.0\linewidth}{@{}l*{4}X@{}}
\toprule
Method           & P    & R    & F    & FPS \\ \midrule
TextSnake~\cite{long2018textsnake}      & 82.7 & 74.5 & 78.4 & -   \\ 
ATRR~\cite{atrr}            & 80.9 & 76.2 & 78.5 & -   \\ 
MTS~\cite{lyu2018mask} & 82.5 & 75.6 & 78.6 & -   \\ 
TextField~\cite{xu2019textfield}        & 81.2 & 79.9 & 80.6 & -   \\ 
LOMO~\cite{lomo}*        & 87.6 & 79.3 & 83.3 & -   \\ 
CRAFT~\cite{craft}           & 87.6 & 79.9 & 83.6 & -   \\ 
CSE~\cite{cse}              & 81.4 & 79.1 & 80.2 & -   \\ 
PSE-1s~\cite{wang2019shape}        & 84.0   & 78.0   & 80.9 & 3.9 \\ 
\midrule  
DB-ResNet-18 (800)   & \textbf{88.3} & 77.9 & 82.8 & \textbf{50}  \\ 
DB-ResNet-50 (800)  & 87.1 & \textbf{82.5} & \textbf{84.7} & 32  \\ \bottomrule
\end{tabularx}
\label{tab:totaltext}
\end{table}

\begin{table}[!ht]
\centering
\caption{Detection results on CTW1500. The methods with ``*" are collected from~\cite{ctw1500}. The values in the bracket mean the height of the input images.}
\begin{tabularx}{1.0\linewidth}{@{}l*{4}X@{}}
\toprule
Method        & P             & R             & F             & FPS         \\ \midrule
CTPN*          & 60.4          & 53.8          & 56.9          & 7.14        \\ 
EAST*          & 78.7          & 49.1          & 60.4          & 21.2        \\ 
SegLink*       & 42.3            & 40.0            & 40.8            & 10.7         \\ 
TextSnake~\cite{long2018textsnake}     & 67.9          & \textbf{85.3}          & 75.6          & 1.1         \\ 
TLOC~\cite{ctw1500}     & 77.4            & 69.8          & 73.4          & 13.3           \\ 
PSE-1s~\cite{wang2019shape}    & 84.8         & 79.7          & 82.2          & 3.9         \\ 
SAE~\cite{tian2019learning}       & 82.7          & 77.8          & 80.1    & 3  \\ 
\midrule  
Ours-ResNet18 (1024) & 84.8          & 77.5          & 81.0          & \textbf{55} \\ 
Ours-ResNet50 (1024) & \textbf{86.9} & 80.2 & \textbf{83.4} & 22          \\ \bottomrule
\end{tabularx}
\label{tab:CTW}
\end{table}

\minisection{Curved text detection}
We prove the shape robustness of our method on two curved text benchmarks (Total-Text and CTW1500). As shown in Tab.~\ref{tab:totaltext} and Tab.~\ref{tab:CTW}, our method achieves state-of-the-art performance both on accuracy and speed.
Specifically, ``DB-ResNet-50" outperforms the previous state-of-the-art method by $1.1\%$ and $1.2\%$ on the Total-Text and the CTW1500 dataset.
``DB-ResNet-50" runs faster than all previous method and the speed can be further improved by using a ResNet-18 backbone, with a small performance drop.
Compared to the recent segmentation-based detector~\cite{wang2019shape}, which runs $3.9$ FPS on Total-Text, ``DB-ResNet-50 (800)" is $8.2$ times faster and ``DB-ResNet-18 (800)" is $12.8$ times faster.

\minisection{Multi-oriented text detection}
The ICDAR 2015 dataset is a multi-oriented text dataset that contains lots of small and low-resolution text instances.
In Tab.~\ref{tab:ic15}, we can see that ``DB-ResNet-50 (1152)" achieves the state-of-the-art performance on accuracy. 
Compared to the previous fastest method~\cite{east}, ``DB-ResNet-50 (736)" outperforms it by $7.2\%$ on accuracy and runs twice faster. For ``DB-ResNet-18 (736)", the speed can be $48$ fps when ResNet-18 is applied to the backbone, with an f-measure of $82.3$.
\begin{table}[ht]
\centering
\caption{Detection results on the ICDAR 2015 dataset. The values in the bracket mean the height of the input images. ``TB'' and ``PSE'' are short for TextBoxes++ and PSENet.}
\begin{tabularx}{1.0\linewidth}{@{}l*{4}X@{}}
\toprule
Method        & P    & R    & F    & FPS  \\ \midrule
CTPN~\cite{eccv/TianHHH016}          & 74.2 & 51.6 & 60.9 & 7.1  \\ 
EAST~\cite{east}          & 83.6 & 73.5 & 78.2 & 13.2 \\ 
SSTD~\cite{sstd}          & 80.2 & 73.9 & 76.9 & 7.7  \\ 
WordSup~\cite{hu2017wordsup}       & 79.3 & 77   & 78.2 & -  \\ 
Corner~\cite{lyu2018multi}    & \textbf{94.1} & 70.7 & 80.7 & 3.6  \\ 
TB~\cite{TextBoxes++}   & 87.2 & 76.7 & 81.7 & 11.6 \\ 
RRD~\cite{liao2018rotation}           & 85.6 & 79   & 82.2 & 6.5  \\ 
MCN~\cite{mcn}           & 72   & 80   & 76   & -  \\ 
TextSnake~\cite{long2018textsnake}     & 84.9 & 80.4 & 82.6 & 1.1  \\ 
PSE-1s~\cite{wang2019shape}    & 86.9 & 84.5 & 85.7 & 1.6  \\ 
SPCNet~\cite{spc}       & 88.7 & \textbf{85.8} & 87.2 & -  \\ 
LOMO~\cite{lomo}      & 91.3 & 83.5 & 87.2 & -  \\ 
CRAFT~\cite{craft}      & 89.8 & 84.3 & 86.9 & -  \\ 
SAE(720)~\cite{tian2019learning}  & 85.1  & 84.5  & 84.8  & 3       \\ 
SAE(990)~\cite{tian2019learning}  & 88.3   & 85.0  & 86.6  & -       \\ 
\midrule  
DB-ResNet-18 (736) & 86.8 & 78.4 & 82.3 & \textbf{48}   \\ 
DB-ResNet-50 (736) & 88.2   & 82.7 & 85.4 & 26   \\ 
DB-ResNet-50 (1152) & 91.8   & 83.2 & \textbf{87.3} & 12   \\ \bottomrule
\end{tabularx}
\label{tab:ic15}
\end{table}

\minisection{Multi-language text detection}
Our method is robust on multi-language text detection. As shown in Tab.~\ref{tab:td500} and Tab.~\ref{tab:mlt}, ``DB-ResNet-50" is superior to previous methods on accuracy and speed. For the accuracy, ``DB-ResNet-50" surpasses the previous state-of-the-art method by $1.9\%$ and $3.8\%$ on the MSRA-TD500 and MLT-2017 dataset respectively. For the speed, ``DB-ResNet-50" is $3.2$ times faster than the previous fastest method~\cite{liao2018rotation} on the MSRA-TD500 dataset.
With a light-weight backbone, ``DB-ResNet-18 (736)" achieves a comparative accuracy compared to the previous state-of-the-art method~\cite{mcn} (82.8 vs 83.0) and runs at 62 FPS, which is $6.2$ times faster than the previous fastest method~\cite{liao2018rotation}, on the MSRA-TD500. The speed can be further accelerated to 82 FPS (``ResNet-18 (512)") by decreasing the input size.

\begin{table}[ht]
\centering
\caption{Detection results on the MSRA-TD500 dataset. The values in the bracket mean the height of the input images.}
\begin{tabularx}{1.0\linewidth}{@{}l*{4}X@{}}
\toprule
Method        & P             & R             & F             & FPS         \\ \midrule
\cite{he2016text}          & 71            & 61            & 69            & -         \\ 
DeepReg~\cite{deepdirect}       & 77            & 70            & 74            & 1.1         \\ 
RRPN~\cite{rrpn}          & 82            & 68            & 74            & -         \\ 
RRD~\cite{liao2018rotation}           & 87            & 73            & 79            & 10          \\ 
MCN~\cite{mcn}          & 88            & 79            & 83            & -         \\ 
PixelLink~\cite{deng2018pixellink}     & 83            & 73.2          & 77.8          & 3           \\ 
Corner~\cite{lyu2018multi}    & 87.6          & 76.2          & 81.5          & 5.7         \\ 
TextSnake~\cite{long2018textsnake}     & 83.2          & 73.9          & 78.3          & 1.1         \\ 
\cite{xue2018accurate}         & 83.0          & 77.4          & 80.1          & -       \\ 
\cite{MSR}     & 87.4          & 76.7          & 81.7          & -       \\   
CRAFT~\cite{craft}         & 88.2          & 78.2          & 82.9          & 8.6       \\ 
SAE~\cite{tian2019learning}        & 84.2          & 81.7          & 82.9          & -       \\ 
\midrule  
DB-ResNet-18 (512) & 85.7          & 73.2          & 79.0          & \textbf{82} \\ 
DB-ResNet-18 (736) & 90.4          & 76.3          & 82.8          & 62 \\ 
DB-ResNet-50 (736) & \textbf{91.5} & \textbf{79.2} & \textbf{84.9} & 32          \\  \bottomrule
\end{tabularx}
\label{tab:td500}
\end{table}

\begin{table}[!ht]
\centering
\caption{Detection results on the MLT-2017 dataset. Methods with ``*" are collected from~\cite{lyu2018multi}. The images in the MLT-2017 dataset are re-sized to $768 \times 1024$ in our method. ``PSE'' is short for PSENet.}
\begin{tabularx}{1.0\linewidth}{@{}l*{4}X@{}}
\toprule
Method        & P             & R             & F             & FPS         \\ \midrule
SARI\_FDU\_RRPN\_V1*       & 71.2            & 55.5            & 62.4            & -         \\ 
Sensetime\_OCR*       & 56.9            & \textbf{69.4}            & 62.6            & -         \\ 
SCUT\_DLVlab1*     & 80.3            & 54.5         & 65.0          & -           \\ 
e2e\_ctc01\_multi\_scale*    & 79.8          & 61.2          & 69.3          & -         \\ 
Corner~\cite{lyu2018multi}          & 83.8          & 55.6          & 66.8          & -        \\ 
PSE~\cite{wang2019shape}          & 73.8            & 68.2            & 70.9            & -         \\ 
\midrule  
DB-ResNet-18 & 81.9          & 63.8          & 71.7          & \textbf{41} \\ 
DB-ResNet-50 & \textbf{83.1} & 67.9 & \textbf{74.7} & 19          \\ \bottomrule
\end{tabularx}
\label{tab:mlt}
\end{table}

\subsection{Limitation}
One limitation of our method is that it can not deal with cases ``text inside text", which means that a text instance is inside another text instance.
Although the shrunk text regions are helpful to the cases that the text instance is not in the center region of another text instance, it fails when the text instance exactly locates in the center region of another text instance. This is a common limitation for segmentation-based scene text detectors.

\section{Conclusion}
In this paper, we have presented a novel framework for detecting arbitrary-shape scene text, which includes the proposed differentiable binarization process (DB) in a segmentation network. The experiments have verified that our method (ResNet-50 backbone) consistently outperforms the state-the-the-art methods on five standard scene text benchmarks, in terms of speed and accuracy. In particular, even with a lightweight backbone (ResNet-18), our method can achieve competitive performance on all the testing datasets with real-time inference speed. In the future, we are interested in extending our method for end-to-end text spotting.

\section{Acknowledgments}
This work was supported by National Key R\&D Program of China (No. 2018YFB1004600), to Dr. Xiang Bai by the National Program for Support of Top-notch Young Professionals and the Program for HUST Academic Frontier Youth Team 2017QYTD08. 

{\fontsize{9.0pt}{10.0pt} \selectfont
\bibliographystyle{aaai}
\bibliography{reference}
}
\end{document}